\documentclass{article}

\usepackage{graphicx} 
\usepackage{float}
\usepackage{booktabs} 
\usepackage{threeparttable}
\usepackage{multirow}
\usepackage{array}
\usepackage{booktabs}
\usepackage{subfig}
\usepackage{amsmath}
\usepackage{bm}
\usepackage{amsfonts}
\usepackage{natbib}
\usepackage{todonotes}
\usepackage{pgfplots}
\usepackage{tikz}
\usepackage{caption}
\usepackage{paralist}
\usepgfplotslibrary{groupplots}

\usepackage{hyperref}

\usepackage[accepted]{icml2020}

\icmltitlerunning{Graph Highway Networks}

\begin{document}

\twocolumn[
\icmltitle{Graph Highway Networks}




\begin{icmlauthorlist}
\icmlauthor{Xin Xin}{gla}
\icmlauthor{Alexandros Karatzoglou}{goo}
\icmlauthor{Ioannis Arapakis}{tel}
\icmlauthor{Joemon M Jose}{gla}
\end{icmlauthorlist}

\icmlaffiliation{gla}{School of Computing Science, University of Glasgow, Scotland, UK}
\icmlaffiliation{goo}{Google Research,London,UK}
\icmlaffiliation{tel}{Telefonica Research, Barcelon, Spain}


\icmlkeywords{Machine Learning, ICML}

\vskip 0.3in
]



\printAffiliationsAndNotice{This work is done when the author was taking an internship in Telefonica Research.}  

\begin{abstract}
Graph Convolution Networks (GCN) are widely used in learning graph representations due to their effectiveness and efficiency. 
However, they suffer from the notorious over-smoothing problem, in which the learned representations of densely connected nodes converge to alike vectors when many ($>3$) graph convolutional layers are stacked. 
In this paper, we argue that the \em re-normalization trick \em used in GCN leads to overly homogeneous information propagation, which is the source of over-smoothing. 
To address this problem, we propose \em Graph Highway Networks \em (GHNet) which utilize gating units to automatically balance the trade-off between \em homogeneity \em and \em heterogeneity \em in the GCN learning process.
The gating units serve as direct highways to maintain  heterogeneous information from the node itself after feature propagation. 
This design enables GHNet to achieve much larger receptive fields per node without over-smoothing and thus access to more of the graph connectivity information.
Experimental results on benchmark datasets demonstrate the superior performance of GHNet over GCN and related models. 
Code will be open-sourced. 
\end{abstract}
\section{Introduction}
Learning dense and low-dimensional node representations from graph-structured data has become the keystone in many practical applications, such as node classification \citep{kipf2016GCN}, protein interface prediction \citep{fout2017protein}, recommendation \citep{berg2017graphmatrixcompletion} and knowledge graphs \citep{GCN_KG}. 
To improve the representation quality,  recent efforts have been focused on adapting well-established deep learning architectures to graph data \citep{cai2018gcn_survey,chen2018fastgcn}. 
From this perspective, one of the most successful models is Graph Convolution Networks (GCN) \citep{kipf2016GCN}, reaching state-of-the-art performance with high efficiency.

\begin{figure*}
    \centering
    \subfloat[1-layer]{
    \label{gcn_cora_1}
    \includegraphics[width=0.19\textwidth]{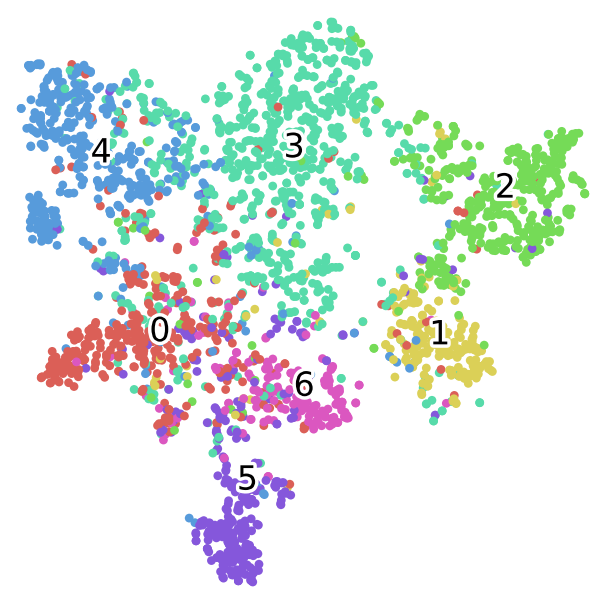}}
    \subfloat[2-layer]{%
    \label{gcn_cora_2}
    \includegraphics[width=0.19\textwidth]{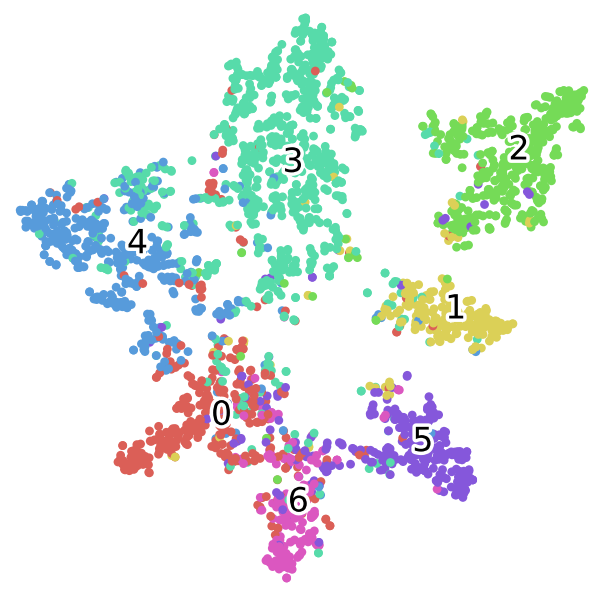}}
    \subfloat[3-layer]{%
    \label{gcn_cora_3}
    \includegraphics[width=0.19\textwidth]{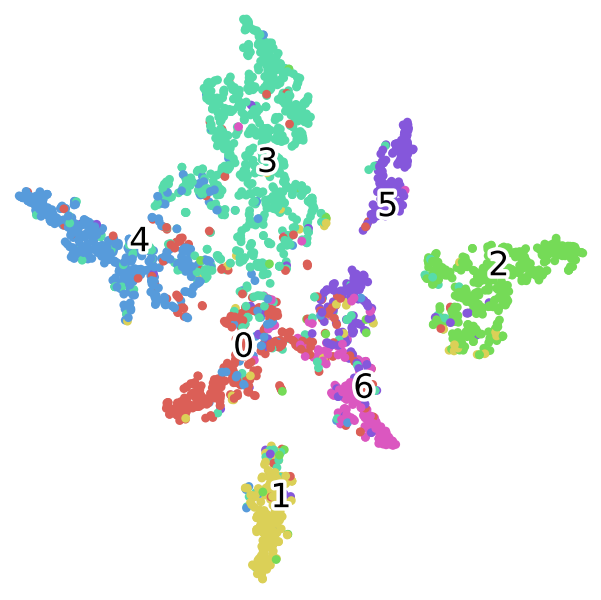}}
    \subfloat[4-layer]{%
    \label{gcn_cora_4}
    \includegraphics[width=0.19\textwidth]{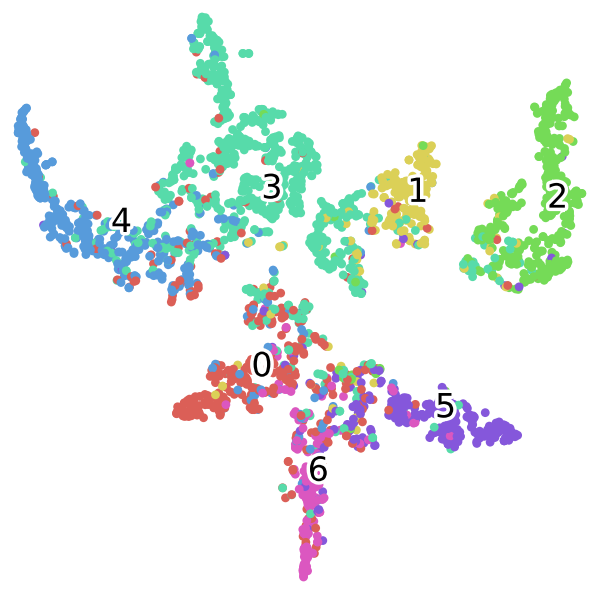}}
    \subfloat[5-layer]{%
    \label{gcn_cora_5}
    \includegraphics[width=0.19\textwidth]{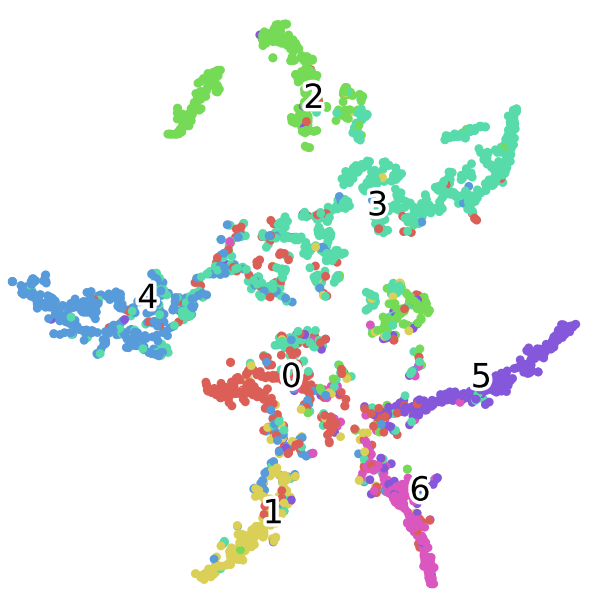}}
    \caption{Node embedding visualization using t-SNE \citep{maaten2008tsne} on Cora dataset with different layers of GCN. The embedding vectors of densely connected nodes tend to converge to the same point and become more indistinguishable with increase of the depth.}
    \label{fig:over-smoothing}
\end{figure*}

While GCN are computationally elegant and effective, one of their main limitations is the depth problem. In original GCN, peak performance is obtained with relatively shallow structures (e.g., 2 or 3 layers), while increasing the depth typically results in dramatic performance degradation \citep{kipf2016GCN}. 
However, using shallow GCN limits the size of the receptive field, which is sub-optimal for feature propagation in sparse data \citep{li2018deeper_insight}.
Although similar depth problems have also been observed in conventional deep learning models \citep{he2016deep_residual_learning}, the main reason for the performance degradation in the graph convolution domain is the ``over-smoothing" problem. This refers to a phenomenon whereby the learned representations of densely connected nodes from GCN tend to converge to an alike vector when the network becomes deeper (see Figure \ref{fig:over-smoothing} for an illustration).

In this paper, we investigate the design of GCN and argue that there is a trade-off between homogeneity and heterogeneity in the learning process. For clarity, we define homogeneity as the property whereby the node representations become a mixture of the representations of the connected nodes (hence similar/homogenized) while heterogeneity as the preservation of the node's original features. We argue that the re-normalization  trick used in GCN leads to overly homogeneous information propagation and thus results into over-smoothing. 

To address this problem, we propose \em Graph Highway Networks \em (GHNet) which utilize learnable gating units to automatically balance the trade-off. 
During each convolution block, the homogeneous information is learned through $k$-hop feature propagation\footnote{Here we remove the GCN self-loop because we propagate the node's own features through a more explicit way.} and the heterogeneous information comes from the node's own features. Subsequently, an element-wise gating function is learned and the output hidden representation is the gated sum of these two parts. The design of GHNet provides two benefits. First, each convolution block performs $k$-hop feature propagation. This enables GHNet to achieve a much larger receptive field per node with only a small number of blocks and thus with fewer parameters, which helps to alleviate potential over-fitting problems. Second, the gating units automatically balance the trade-off between homogeneity and heterogeneity in the learning process. This allows the node to receive information from a much larger receptive field while it also preserves enough of the original node features. The main contributions of this paper are summarized as follows:
\begin{itemize}
    \item We investigate the homogeneity/heterogeneity trade-off of GCN and provide new insights on the over-smoothing problem.
    \item We propose GHNet which utilize gating units to automatically learn the balance between homogeneity and heterogeneity. We design different variants of GHNet that maintain the node feature distinctiveness after feature propagation.
    \item We conduct experiments on benchmark datasets to validate the proposed models. Experimental results show that GHNet outperforms GCN and other related models. 
\end{itemize}

\section{Graph Convolutional Networks}
\label{sec:GCN investigation}
\subsection{Recap}
GCN were proposed by \citet{kipf2016GCN} for semi-supervised node classification. In this setting, GCN are applied to a graph with partial-labeled nodes as input and generate label predictions for other nodes. A graph is defined as $\mathcal{G}=(\mathcal{V},\mathbf{A})$, where $\mathcal{V}$ represents the node set $\left\{ v_1,v_2,...,v_n \right\}$ and $\mathbf{A}\in \mathbb{R}^{n\times n}$ is the adjacent matrix with $a_{ij}$ denoting the $(i,j)$-th entry. $a_{ij}=1$ indicates the existence of an edge between node $i$ and node $j$, otherwise $a_{ij}=0$. Each node $v_i$ in the graph has a corresponding feature vector $\mathbf{x}_{i} \in \mathbb{R}^d$ and the entire input feature matrix is denoted as $\mathbf{X}=[\mathbf{x}_1,\mathbf{x}_2,...,\mathbf{x}_n]\in\mathbb{R}^{n\times d}$.
GCN take the above information as the input and generate low-dimensional dense node representations, which are eventually feed into a softmax function to perform classification.

The initial spectral convolution derived in GCN is formulated as
\begin{equation}
	\label{equation:initial_gcn}
	g_{\theta}\star \mathbf{x}_i=\theta(\mathbf{I}+\mathbf{D}^{-\frac{1}{2}}\mathbf{A}\mathbf{D}^{-\frac{1}{2}})\mathbf{x}_i ,
\end{equation}
where $g_{\theta}$ is the spectral filter, $\star$ denotes the convolution operator, $\mathbf{I}$ is an identity matrix and $\mathbf{D}$ is the diagonal degree matrix of $\mathbf{A}$. Due to the fact that $\mathbf{I}+\mathbf{D}^{-\frac{1}{2}}\mathbf{A}\mathbf{D}^{-\frac{1}{2}}$  has eigenvalues in
the range [0, 2], repeated application of this operator can therefore lead to numerical instabilities and gradient issues \citep{wu2019SGC,kipf2016GCN}. As a result, \citet{kipf2016GCN} proposed the \em re-normalization trick \em as 
\begin{equation}
	\label{equation:renormalization}
	\mathbf{I}+\mathbf{D}^{-\frac{1}{2}}\mathbf{A}\mathbf{D}^{-\frac{1}{2}}\rightarrow
	\widetilde{\mathbf{D}}^{-\frac{1}{2}}
	\widetilde{\mathbf{A}}
	\widetilde{\mathbf{D}}^{-\frac{1}{2}},
\end{equation}
where $\widetilde{\mathbf{A}}=\mathbf{A}+\mathbf{I}$ and $\widetilde{\mathbf{D}}$ is the corresponding diagonal matrix of $\widetilde{\mathbf{A}}$. Combining the derived filter with a neural network, the final formulation of GCN becomes: 
\begin{equation}
	\label{equation:gcn_final}
	\mathbf{H}^{(l)}=\begin{cases}\mathbf{X} & l = 0\\
 \sigma(\widetilde{\mathbf{D}}^{-\frac{1}{2}}\widetilde{\mathbf{A}}\widetilde{\mathbf{D}}^{-\frac{1}{2}}\mathbf{H}^{(l-1)}\mathbf{\Theta}^{(l-1)})& l \geq 1,\end{cases}
\end{equation}
where $\mathbf{H}^{(l)}$ denotes the output of the $l$-th layer, $\sigma$ is the activation function and $\mathbf{\Theta}^{(l-1)}$ is the transition matrix.
\subsection{Homogeneity Heterogeneity Trade-off}
We can see that each GCN layer contains two stages: feature propagation and nonlinear transition. To keep it simple, we only focus on the feature propagation process.

Let $\overline{\mathbf{h}}_i^{(l)}$ denote the intermediate representations of $v_i$ in the $l$-th layer after feature propagation. The propagation rule defined by the filter in Eq.(\ref{equation:initial_gcn}) can be written as
\begin{equation}
	\label{equation:hetergeneous_feature_propagation}
	\overline{\mathbf{h}}_i^{(l)} =\mathbf{h}_i^{(l-1)}
	+\sum_{j=1}^n\frac{a_{ij}}{\sqrt{d_id_j}}\mathbf{h}_j^{(l-1)},
\end{equation}
where $\mathbf{h}_i^{(l)}$ denotes the corresponding representations of node $v_i$ in 	$\mathbf{H}^{(l)}$ and the $d_i$ is the degree of $v_i$ as $d_i=\sum_ja_{ij}$.
We note that compared with the coefficient of a random neighbor (i.e., $a_{ij}/\sqrt{d_id_j}$), the weight of the node itself (i.e., 1) is much larger. This means that the node's original feature account for the highest importance among all nodes during the propagation process. This helps to preserve the node's distinct feature information. However, this big difference between convolution weights also induces instability in the learning process, especially when the filters are applied multiple times \citep{kipf2016GCN,wu2019SGC}. As a result, we can claim that the filters defined in Eq.(\ref{equation:initial_gcn}) lead to overly heterogeneous feature propagation.

To enhance stability, the re-normalization trick described in Eq.(\ref{equation:renormalization}) is introduced and the propagation rule after normalization can be written as 
\begin{equation}
	\label{equation:homogenerous_feature_propagation}
	\overline{\mathbf{h}}_i^{(l)}\!\!=\!\frac{1}{d_i\!+\!1}\mathbf{h}_i^{(l-1)}
	\!+\!\sum_{j=1}^n\frac{a_{ij}}{\sqrt{(d_i\!+\!1)(d_j\!+\!1)}}\mathbf{h}_j^{(l-1)}.
\end{equation}
The intuition behind Eq.(\ref{equation:homogenerous_feature_propagation}) is that a ``self-loop" is augmented to connect the node to itself so the node now becomes a member of its own neighbourhood. This smooths the convolution weights and thus helps to achieve a more stable learning process \citep{wu2019SGC,li2018deeper_insight}. 
However, it also happens that the node is regarded as a ``normal" neighbour, hence no extra importance is assigned to the node itself during the aggregation between neighboring nodes. 
As a result, after repeated feature mixtures, the node fails to preserve its own distinct features and the representations of densely connected nodes become more and more similar, this is also known as over-smoothing. Contrary to Eq.(\ref{equation:hetergeneous_feature_propagation}), the propagation rule of Eq.(\ref{equation:homogenerous_feature_propagation}) results into overly homogeneous feature propagation.

Generally speaking, there is a trade-off between heterogeneity and homogeneity in the learning process of GCN. 
The re-normalization trick promotes homogeneity, which makes the learning process more stable (similar to regularization) but also increases the risk of over-smoothing. 
While a learning process that preservers the node's distinct features allows the node to receive information from a larger receptive field without homogenizing the representations. However, overly heterogeneous node features lead to instability problems. 
\section{Graph Highway Networks}
To automatically balance homogeneity and heterogeneity in the learning process, and encourage the node to receive information from a large receptive filed without over-smoothing, we propose GHNet, which includes multi-hop feature propagation and gating units.

Based on the discussion in Section \ref{sec:GCN investigation}, we formulate the output of a convolution block in GHNet as the gated sum of two components:
\begin{equation}
	\label{equation:overall_equation_ghnet}
	\mathbf{H}^{(l+k)} =\mathbf{T}^{(l)}\odot \mathbf{F}_{hom}+(1-\mathbf{T}^{(l)})\odot \mathbf{F}_{het} ,
\end{equation}
where the superscript $(l+k)$ indicates that we perform $k$-hop feature propagation in this block, $\odot$ denotes element-wise product, $\mathbf{F}_{hom}$ is the homogeneous representation while $\mathbf{F}_{het}$ is the heterogeneous component. $\mathbf{T}^{(l)}$ is the output of the gating function, which is formulated as 
\begin{equation}
	\label{equation:gating_function}
	\mathbf{T}^{(l)} =\delta(\mathbf{W}_T^{(l)}\mathbf{H}^{(l)}+b^{(l)}).
\end{equation}
$\delta(\cdot)$ is the sigmoid function. $\mathbf{W}_T^{(l)}$ and $b^{(l)}$ are learnable parameters. In the following subsections, we will introduce the detail to model $\mathbf{F}_{hom}$ and $\mathbf{F}_{het}$.
\begin{figure*}
	\centering
    \includegraphics[width=0.8\textwidth]{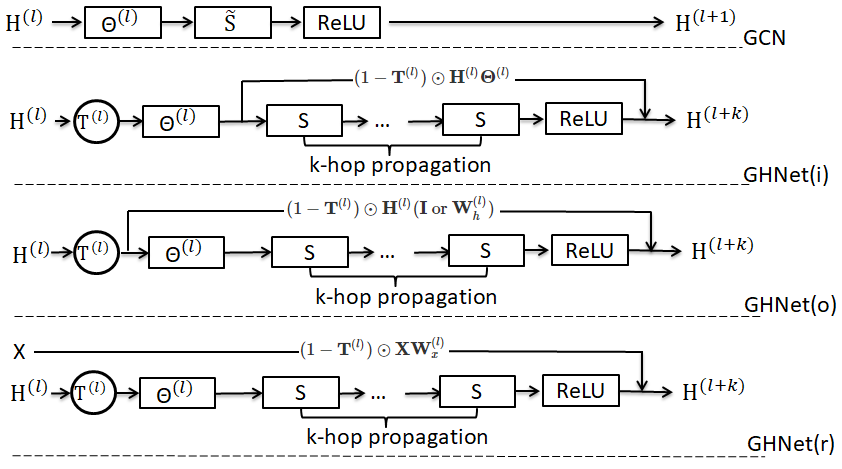}
    \caption{Structures of a convolution block in GCN and GHNet. $\widetilde{\mathbf{S}}$ and $\mathbf{S}$ denote the corresponding filters with $\widetilde{\mathbf{S}}=	\widetilde{\mathbf{D}}^{-\frac{1}{2}}
	\widetilde{\mathbf{A}}
	\widetilde{\mathbf{D}}^{-\frac{1}{2}}$ and  $\mathbf{S}=\mathbf{D}^{-\frac{1}{2}}\mathbf{A}\mathbf{D}^{-\frac{1}{2}}$.}
    \label{fig:model_structures}
\end{figure*}

\subsection{Multi-hop Feature Propagation}
The most important ingredient, which distinguishes GCN from conventional neural networks (e.g., multi-layer perceptrons), is the aggregation between neighbourhoods. This enables the revealed knowledge to flow through the graph and benefit downstream tasks. A large receptive field can help to guarantee adequate information flows and thus improve learning performance, especially in sparse labeled settings \citep{li2018deeper_insight}.

However, as demonstrated by  Eq.(\ref{equation:gcn_final}), GCN perform a matrix transformation through a layer specific parameter $\mathbf{\Theta}^{(l)}$ after every hop of propagation. As a result, increasing the receptive field will simultaneously increase the number of parameters and, consequently, the risk of over-fitting.

To address the above issues, here we adopt a ``batched" operation which means that we perform $k$-hop propagation in a single block. Inspired by the observation that one-hop propagation corresponds to a convolution with the spectral filter, we define the homogeneous representation after $k$-hop propagation as 
\begin{equation}
	\label{equation:F_hom}
	\mathbf{F}_{hom}=\sigma((\mathbf{D}^{-\frac{1}{2}}\mathbf{A}\mathbf{D}^{-\frac{1}{2}})^k\mathbf{H}^{(l)}\mathbf{\Theta}^{(l)}).
\end{equation}
This design enables GHNet to achieve a large receptive field with only a small number of convolution blocks and thus with fewer parameters.

Note that here we use $\mathbf{A}$ and $\mathbf{D}$ to perform propagation without the “self-loop”.
Because the node's own feature information will be modeled with $\mathbf{F}_{het}$.

\subsection{Heterogeneous Information Infusion}
The keystone in alleviating over-smoothing is to introduce the node's own features after propagation. In this subsection, we propose three variants to model the (heterogeneous) node feature information.
\subsubsection{Inner infusion}
This variant models the node's own features inside the block, which means we use the node's representation after the matrix transformation but before the convolution, as the heterogeneous representation: 
\begin{equation}
	\label{equation:inner_infusion}
	\mathbf{F}_{het}=\mathbf{H}^{(l)}\mathbf{\Theta}^{(l)}.
\end{equation}
Under this setting, the model structure can be demonstrated as GHNet(i) in Figure \ref{fig:model_structures}. The advantage of inner infusion is that $\mathbf{F}_{het}$ and $\mathbf{F}_{hom}$ are in the same latent space with the same dimensionality. As a result, no dimension adjustment is  required. 
\subsubsection{Outer infusion}
This variant targets at representing the nodes own features outside the whole block and directly infuse $\mathbf{H}^{(l)}$ into the output. However, when the dimensions of $\mathbf{H}^{(l)}$ and $\mathbf{H}^{(l+k)}$ are different, an adjustment operation needs to be performed. Under this case, the formulation of $\mathbf{F}_{het}$ is shown as
\begin{equation}
	\label{equation:outer_infusion}
	\mathbf{F}_{het}=\begin{cases}\mathbf{H}^{(l)} & I(\mathbf{H}^{(l)},\mathbf{H}^{(l+k)}) = 1\\\mathbf{H}^{(l)}\mathbf{W}_h^{(l)} & I(\mathbf{H}^{(l)},\mathbf{H}^{(l+k)}) = 0,\end{cases}
\end{equation}
where $I(\cdot,\cdot)$ is an identification function with 1 denoting the two inputs have the same dimensionality and 0 otherwise. $\mathbf{W}_h^{(l)}$ is the matrix used to perform dimension projection. The model structure of this variant is demonstrated as GHNet(o) in Figure \ref{fig:model_structures}.
\subsubsection{Raw infusion}
Another solution assumes the heterogeneous information is independent to the current state (i.e., $\mathbf{H}^{(l)}$) but only determined by the raw input features (i.e., $\mathbf{X}$). Under this assumption, the heterogeneous representation is formulated as
\begin{equation}
	\label{equation:raw_infusion}
	\mathbf{F}_{het}=\mathbf{X}\mathbf{W}_{x}^{(l)},
\end{equation}
where $\mathbf{W}_{x}^{(l)}$ is the weight matrix to perform dimension adjustment. The model structure of this variant is shown in Figure \ref{fig:model_structures} as GHNet(r).

To conclude, in this section, we propose three variants of GHNet to perform multi-hop feature propagation and heterogeneous information infusion. The trade-off between the two components is automatically balanced through a learnable gating function.
\section{Experiments}
In this section, we conduct experiments to evaluate the performance of GHNet on the node classification task. In this task, we take a graph with partially labeled nodes as the input and generate label predictions for unlabeled nodes. Classification accuracy is used as the evaluation metric. We aim to answer the following research questions:

\textbf{RQ1:} Compared with GCN and other related models, how does GHNet perform?

\textbf{RQ2:} How do the designs of GHNet affect the model performance, including gating units and multi-hop propagation?

\textbf{RQ3:} Does GHNet help to resolve the over-smoothing problem?
\subsection{Experimental Setting}
\subsubsection{Datasets}
The experiments are conducted on benchmark datasets, including citation networks and knowledge graphs. Dataset statistics are summarized in Table \ref{tbl:datasets}. For the citation network datasets (i.e., Cora, Citeseer and Pubmed), nodes
are documents and edges are citation links. 
For the knowledge graph dataset (i.e., NELL), we apply the same pre-processing steps of \citet{kipf2016GCN}. For all datasets, we use exactly the same data splits as \citet{kipf2016GCN} for training, validation and test without special mention.

\subsubsection{Baselines}
We compare the performance of the proposed three variants of GHNet with the following baselines:
\begin{itemize}
    \item MLP: The standard multi-layer perceptron with softmax as the final layer to perform classification. It serves as a basement comparison for other methods.
	\item GCN: The original graph convolution network which utilizes the self-loop and re-normalization tricks \citep{kipf2016GCN}.
	\item SGC: Simple graph convolution is the fast version of GCN. It removes the non-linear transition and treats feature propagation as a pre-computing process \citep{wu2019SGC}.
	\item JKL: Jumping knowledge network is an ensemble learning approach which combines the hidden representations of different GCN layers \citep{xu2018jumping}.
	\item MixHop: MixHop is a recently proposed approach which learns mix-order neighborhood information through concatenating high-order convolutions \citep{kapoor2019mixhop}.
\end{itemize}

For GCN, SGC and MixHop we use the implementations given by their authors. For JKL, we use our own implementation due to unavailability of the original code. 
\begin{table}
    \centering
    \begin{tabular}{p{1.2cm}<{\centering}p{1cm}<{\centering}p{1cm}<{\centering}p{1.0cm}<{\centering}p{2.0cm}<{\centering}}
        \toprule
        \textbf{Dataset}  & \textbf{Nodes} & \textbf{Edges} & \textbf{Classes} & \textbf{Label rate}  \\
        \midrule
        Cora   &   2,708  &   5,429  &   7  &  0.052   \\
        Citeseer    &   3,327  &   4,732  &   6  &   0.036   \\
        Pubmed    &   19,717  &   44,338  &   3  &   0.003   \\
        NELL     &   65,755  &   266,144  &   210  &  0.010 \\
        \bottomrule
    \end{tabular}
    \caption{Dataset statistics.}
    \label{tbl:datasets}
\end{table}

\subsubsection{Parameter settings}
We train all three variants of GHNet with two convolution blocks in Cora, Citeseer, Pubmed and three convolution blocks in NELL. All models are trained with the Adam \citep{kingma2014adam} optimizer. The learning rate is 0.01 for Cora, Citeseer, Pumbed and 0.02 for the NELL dataset. The number of hops in each block (i.e., $k$) is tuned within \{1,2,3,4,5\}. The weights are initialized using glorot initializer \citep{glorot2010glorot}. The hidden size is set as the same with GCN for a fair comparison, which is 16 in Cora, Citeseer, Pubmed and 64 in NELL. The early-stop strategy and dropout with 0.5 drop-ratio are also introduced to prevent over-fitting. For GCN, SGC and MixHop, we use the exact same settings as described in their papers. For JKL, we use the ensemble approaches described in their original paper (i.e., mean-pooling, max-pooling and LSTM) and just report the highest results. All experiments are conducted 5 times, with a different random seed each time, and the average scores are reported.

\subsection{Performance Comparison (RQ1)}
Table \ref{tbl:performance comparison} shows the performance comparison between all models in terms of classification accuracy. The accuracy of GCN on NELL dataset is 66.0\% according to their original paper. 
However, follow-up work, including ours, has encountered difficulties when reproducing this result\footnote{See https://github.com/tkipf/gcn/issues/14 for detail.}. 63.3\% is the best accuracy that we can reproduce. 
In addition, we find that SGC can't reach an acceptable performance on the NELL dataset. The reason may lie in the fact that SGC removes all non-linearities and thus the model becomes equivalent to a linear regression model. Although it can achieve promising results on small datasets like Cora and Citesser, the models expressiveness isn't sufficient to fit the larger NELL dataset. We are still investigating this problem.

We can see that the best performance on all datasets is achieved by the proposed GHNet. Meanwhile, all the proposed three variants of GHNet achieve better results than the original GCN (except for GHNet(r) in Citeseer, but their performance is a close tie). The reason lies in the fact that gating units in GHNet can automatically balance the trade-off between homogeneity and heterogeneity so the node can receive information from a large receptive field without losing his own features.


Furthermore, we observe that JKL doesn't outperform the original GCN on the citation network datasets (i.e., Cora, Citeseer and Pubmed). We believe the reason is that the authors of JKL utilize different data splits in their paper, which increases the portion of training data. Since JKL is an ensemble learning approach that combines the outputs of different layers, it may require more training instances to attain better performance. However, it still performs better than GCN on the NELL dataset. MixHop also combines the hidden representation of different convolution layers but it enables to learn an adaptive architecture and thus gets a better performance compared with vanilla GCN. 

\begin{table}
    \centering
    \begin{tabular}{p{1.4cm}<{\centering}p{1cm}<{\centering}p{1.2cm}<{\centering}p{1.2cm}<{\centering}p{1.2cm}<{\centering}}
        \toprule
        \textbf{Model} & \textbf{Cora} & \textbf{Citeseer} &\textbf{Pubmed} & \textbf{NELL}\\
        \midrule
        MLP    &   56.1  &   55.3  &   72.1  &   32.4  \\
        GCN&81.5&70.3&79.3&63.3\\
        SGC&81.0&69.8&78.9&35.9\\
        JKL&79.7&69.1&78.2&66.6\\
        MixHop&81.9&70.6&79.7&69.3\\
        \midrule
        GHNet(i)&$\mathbf{82.6}^{\star \star}$&$70.9$&$\mathbf{80.3}^{\star \star}$&$72.5^{\star \star}$\\
        GHNet(o)&81.7&$\mathbf{71.3}^{\star}$&$79.9$&$\mathbf{73.4}^{\star \star}$\\
        GHNet(r)&81.9&70.2&$79.9$&$70.2^{\star}$\\
        \bottomrule
    \end{tabular}
    \caption{Classification accuracy of different models (in percentage). \textbf{Boldface} denotes highest scores. $^{\star}$ and $^{\star \star}$ denote the statistical significance for $p<0.1$ and $p<0.05$ compared with the best baseline, correspondingly.}
    \label{tbl:performance comparison}
\end{table}
\begin{table}
    \centering
    \begin{tabular}{p{1.8cm}<{\centering}p{0.8cm}<{\centering}p{0.8cm}<{\centering}p{0.8cm}<{\centering}p{0.8cm}<{\centering}p{0.8cm}<{\centering}}
        \toprule
        \textbf{Label rate} & \textbf{0.005} & \textbf{0.010} &\textbf{0.015} & \textbf{0.020}& \textbf{0.025}\\
        \midrule
        GCN    &   48.9  &   58.1  &   68.2  &   73.5&  74.0\\
        GHNet(i)&$53.4^{\star\star}$&$68.0^{\star\star}$&$69.9^{\star\star}$&$75.1^{\star\star}$&$75.3^{\star\star}$\\
        GHNet(o)&$55.6^{\star\star}$&$65.3^{\star\star}$&$71.9^{\star\star}$&$74.0^{\star}$&$74.7^{\star}$\\
        GHNet(r)&$53.5^{\star\star}$&$64.2^{\star\star}$&$72.5^{\star\star}$&$74.2^{\star}$&$74.7^{\star}$\\
        \midrule
        &10.8\%&13.3\%&4.7\%&1.3\%&1.2\%\\
        \bottomrule
    \end{tabular}
    \caption{Classification accuracy on Cora with different label rates. The last row indicates average relative improvement of GHNet compared with GCN. $^{\star}$ and $^{\star \star}$ denote statistical significance for $p<0.1$ and $p<0.05$ compared with GCN, correspondingly.}
    \label{tbl:sparse_cora}
\end{table}
We also observe that the improvement over baselines is much larger on the NELL dataset. The reason may be that this dataset is more complex (e.g., including 210 classes) but the label rate is relatively sparse. This lead to a circumstance that a large receptive field is particularly desirable to guarantee adequate information flows. To verify this assumption, we conduct experiments to benchmark the performance of GHNet on one dataset with different sparsity (label rates). 
We randomly select 10\%, 20\%, 30\%, 40\% and 50\% of the training instances\footnote{Correspodning to the label rates of 0.005, 0.010, 0.015, 0.020, 0.025.} in Cora dataset as labeled data and train the models. In Table \ref{tbl:sparse_cora} we can see that the relative improvement of GHNet is much larger when the label rate is sparse. The reported results further confirm the advantage of GHNet when there is only limited training labels.

\begin{table}
    \centering
    \begin{tabular}{p{0.8cm}<{\centering}p{1.6cm}<{\centering}p{1.6cm}<{\centering}p{1.6cm}<{\centering}}
        \toprule
        $t$ & \textbf{inner} & \textbf{outer} &\textbf{raw} \\
        \midrule
        0.1   &   65.5($\pm$1.59) &   67.2($\pm$0.94)  &   62.4($\pm$0.71) \\
        0.3&76.0($\pm$0.55)&76.8($\pm$0.62)&77.0($\pm$0.66)\\
        0.5&81.3($\pm$0.35)&81.4($\pm$0.52)&80.0($\pm$0.47)\\
        0.7&82.1($\pm$0.29)&81.3($\pm$0.59)&81.2($\pm$0.52)\\
        0.9&80.6($\pm$0.42)&80.1($\pm$0.98)&80.9($\pm$0.41)\\
        \midrule
        gate&\textbf{82.6}($\pm$0.35)&\textbf{81.7}($\pm$0.50)&\textbf{81.9}($\pm$0.42)\\
        \bottomrule
    \end{tabular}
    \caption{Classification accuracy on Cora dataset when replacing gating unites with pre-defined scalar $t$.}
    \vspace{-0.5cm}
    \label{tbl:effect_of_gate}
\end{table}
\subsection{Model Investigation (RQ2)}
\subsubsection{Effect of gating units}
In this section, we conduct experiments to demonstrate the effectiveness of the introduced gating units in GHNet. We replace the element-wise gating function with a pre-defined scalar $t$ and keep the other settings fixed. The formulation of $\mathbf{H}^{(l+k)}$ now can be written as $\mathbf{H}^{(l+k)} =t\cdot \mathbf{F}_{hom}+(1-t)\cdot \mathbf{F}_{het}$. Table \ref{tbl:effect_of_gate} shows the result on Cora dataset\footnote{Results on other datasets lead to same observations and are omitted due to the space limitation.}. We can see that GHNet with the learnable gating function achieves the best performance, compared with all ranges of $t$. In fact, the introduced gating function not only balance the importance between $\mathbf{F}_{hom}$ and $\mathbf{F}_{het}$ automatically but also provide dimension-wise adaption, which increases the model fidelity and thus boosts performance. Besides, we can also see that the classification accuracy improves with the increase of $t$ at the beginning and then drops. This verifies our proposition that there is a trade-off between homogeneity and heterogeneity in the convolution procedure. Both overly homogeneous or heterogeneous information propagation will degrade the model performance.

\begin{figure*}
\setlength{\abovecaptionskip}{-0.1cm}
    \centering
    \subfloat[Cora]{
    \label{gate_cora_1}
    \includegraphics[width=0.3\textwidth]{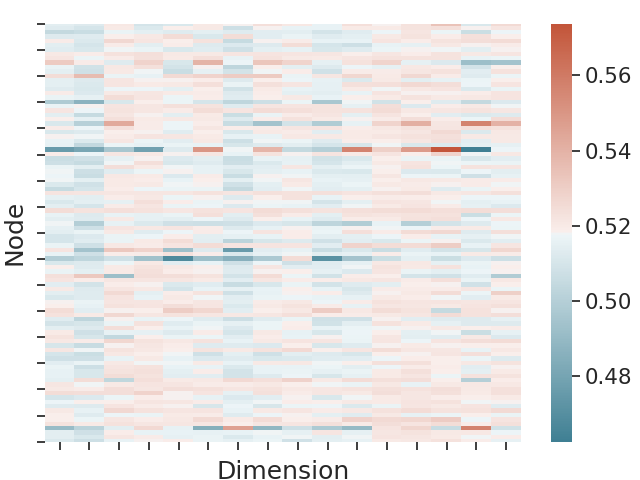}}
    \subfloat[Citeseer]{%
    \label{gate_citeseer_1}
    \includegraphics[width=0.3\textwidth]{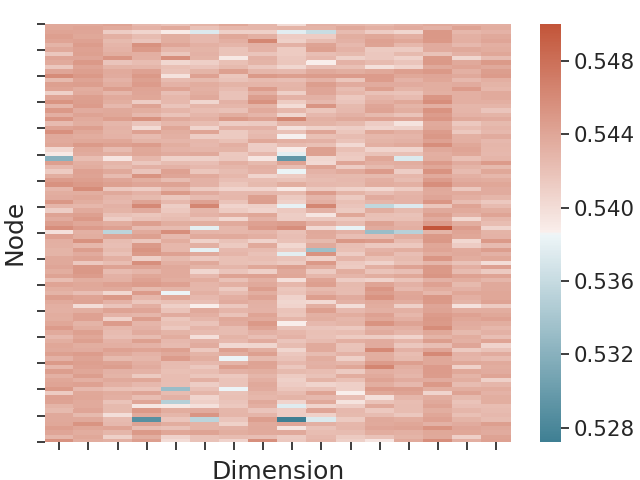}}
    \subfloat[Pubmed]{%
    \label{gate_pubmed_1}
    \includegraphics[width=0.3\textwidth]{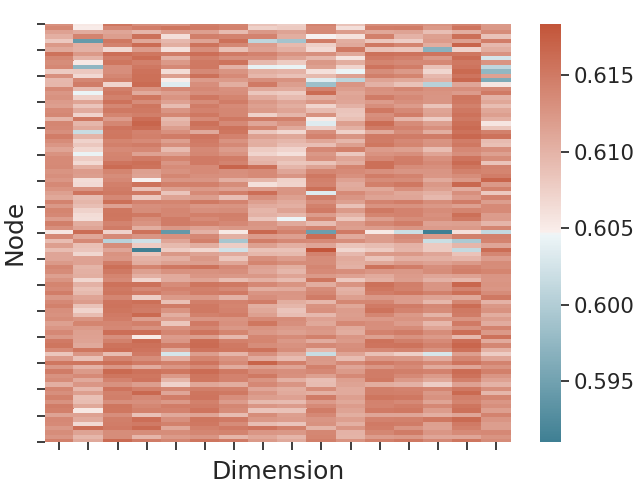}}
    \caption{Visualization of learned gating outputs in the first convoltion block}
    \label{fig:gate_1}
\vspace{-0.6cm}
\end{figure*}

\begin{figure*}
    \centering
    \subfloat[Cora]{
    \label{gate_cora_2}
    \includegraphics[width=0.3\textwidth]{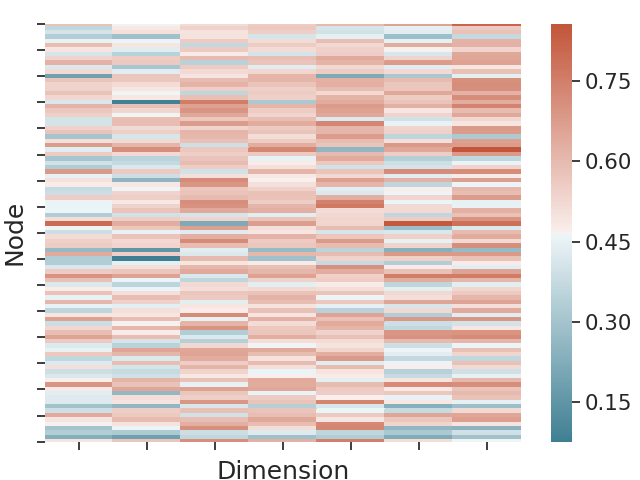}}
    \subfloat[Citeseer]{%
    \label{gate_citeseer_2}
    \includegraphics[width=0.3\textwidth]{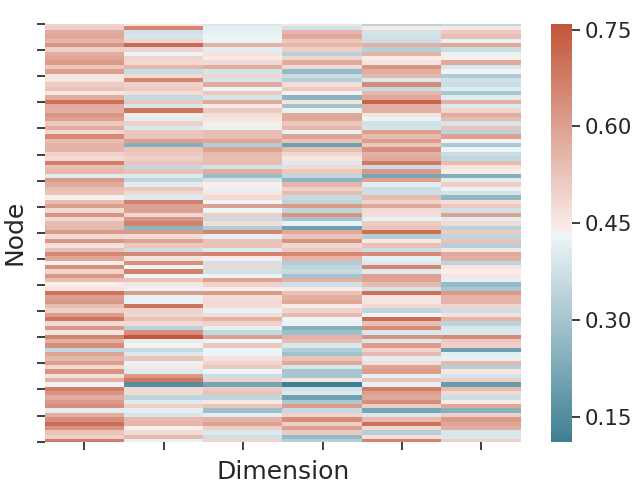}}
    \subfloat[Pubmed]{%
    \label{gate_pubmed_2}
    \includegraphics[width=0.3\textwidth]{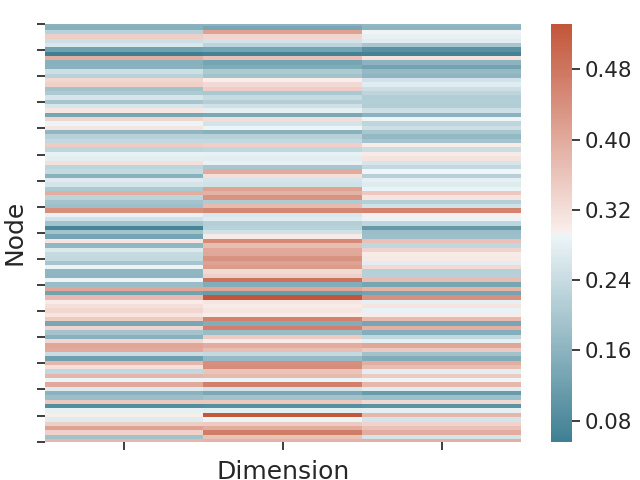}}
    \caption{Visualization of learned gating outputs in the second convolution block}
    \label{fig:gate_2}
\end{figure*}

In addition, we provide a visualization of the learned gating outputs (in the case of GHNet(i)). We randomly choose 100 nodes in Cora, Citeseer, Pubmed and draw the distribution of the corresponding gates. Figure \ref{fig:gate_1} and Figure \ref{fig:gate_2} show the results in the first and second block, respectively. We can see that most gates in the first block fall into a relatively small range between 0.4$\sim$0.6. However, the distribution in the second block is more discrete and spans a much larger range. The reason may be that the first convolution block serves as a fundamental player to capture macro-level patterns in the graph while the second block is more task-oriented and node-specific which aims to increase the model expressiveness and fidelity.

\subsubsection{Effect of multi-hop propagation}
Another design feature of GHNet is multi-hop propagation in one convolution block. In this part, we conduct experiments to show the effect of hops (i.e., $k$). We set $k=1$ in the first convolution block and show the results when tuning $k$ in the other block. Figure \ref{fig:effect_hops} shows the result on Cora, Citeseer and Pubmed\footnote{We don't show the results on NELL because we use three convolution blocks on this dataset.}.
We observe that on all three datasets, the performance is improved when $k$ increases from 1. This confirms the advantage of carrying out multi-hop propagation which can increase the receptive field more efficiently. Specifically, on the Cora dataset, the performance keeps increasing and reaches top performance when $k=5$. However,  we note that on on Citeseer and Pubmed, the performance may also decrease with a $k$ that is too large. The reason may be that a large $k$ will simultaneously increase the risk of introducing noisy information. 


\begin{center}
\begin{table*}
    \begin{tabular}{p{\textwidth}}
        \begin{center}
        \ref{named}
        \pgfplotsset{every axis/.append style={
            height=4.5cm,
            width=6cm,
            xmin=0.85, xmax=5.15,
            xtick={1,2,3,4,5},
            grid=major,
            grid style=dashed,
            tick style={ultra thin},
            tick label style={font=\tiny},
            label style={font=\footnotesize},
            legend style={font=\tiny},
            title style={font=\small}
            }
        }
        
        \tikzset{every mark/.append style={scale=0.8}}
        \begin{tikzpicture}[trim axis left]
        \begin{axis}[
            legend columns=-1,
            legend entries={GHNet(i),GHNet(o), GHNet(r)},
            legend to name=named,
            xlabel={(a) Cora},
            every axis x label/.style={at={(0.5,-0.2)}},
            ylabel={Accuracy},
            ymin=0.79, ymax=0.83,
            ytick={0.79,0.8,0.81,0.82,0.83},
        ]
        \addplot[
            color=blue,
            mark=otimes*,
            ]
            coordinates {
            (1,0.802)(2,0.811)(3,0.822)(4,0.822)(5,0.825)
            };
        \addplot[
            color=red,
            mark=square*,
            ]
            coordinates {
            (1,0.798)(2,0.801)(3,0.803)(4,0.812)(5,0.819)
            };
        \addplot[
            color=teal,
            mark=diamond*,
            ]
            coordinates {
            (1,0.799)(2,0.803)(3,0.814)(4,0.811)(5,0.817)
            };
        \end{axis}
        \end{tikzpicture}
        \hspace{0.25cm}
        \begin{tikzpicture}
        \begin{axis}[
            xlabel={(b) Citeseer},
            every axis x label/.style={at={(0.5,-0.2)}},
            ymin=0.65, ymax=0.71,
            ytick={0.65,0.67,0.69,0.71},
        ]
        \addplot[
            color=blue,
            mark=otimes*,
            ]
            coordinates {
            (1,0.658)(2,0.671)(3,0.698)(4,0.692)(5,0.689)
            };
        \addplot[
            color=red,
            mark=square*,
            ]
            coordinates {
            (1,0.68)(2,0.705)(3,0.703)(4,0.7)(5,0.698)
            };
        \addplot[
            color=teal,
            mark=diamond*,
            ]
            coordinates {
            (1,0.676)(2,0.69)(3,0.7)(4,0.697)(5,0.699)
            };
        \end{axis}
        \end{tikzpicture}
        \hspace{0.25cm}
        \begin{tikzpicture}[trim axis right]
        \begin{axis}[
            every axis x label/.style={at={(0.5,-0.2)}},
            xlabel={(c) Pubmed},
            ymin=0.76, ymax=0.80,
            ytick={0.76,0.77,0.78,0.79,0.80},
        ]
        \addplot[
            color=blue,
            mark=otimes*,
            ]
            coordinates {
            (1,0.774)(2,0.79)(3,0.792)(4,0.79)(5,0.791)
            };
        \addplot[
            color=red,
            mark=square*,
            ]
            coordinates {
            (1,0.767)(2,0.771)(3,0.788)(4,0.797)(5,0.794)
            };
        \addplot[
            color=teal,
            mark=diamond*,
            ]
            coordinates {
            (1,0.787)(2,0.79)(3,0.796)(4,0.798)(5,0.8)
            };
        \end{axis}
        \end{tikzpicture}
        
        \end{center}
    \end{tabular}
    \vspace{-0.9cm}
    
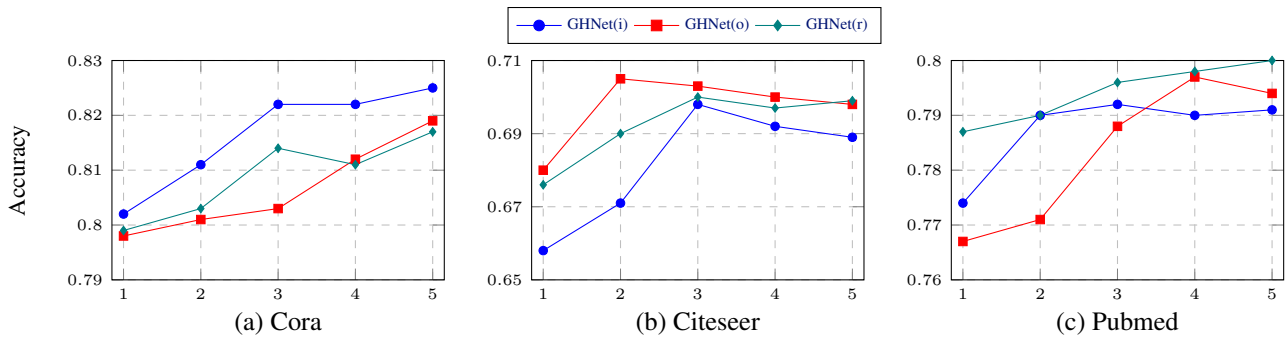
\captionof{figure}{Effect of hops (i.e., $k$) on Cora, Citeseer and Pubmed.}
    \label{fig:effect_hops}
\end{table*}
\end{center}

\subsection{Embedding Visualization (RQ3)}
\begin{figure*}
    \centering
    \subfloat[5-layer GCN]{
    \label{gcn_cora_5_for_comparison}
    \includegraphics[width=0.23\textwidth]{gcn_cora_5.png}}
    \subfloat[GHNet(i)]{%
    \label{inner_cora}
    \includegraphics[width=0.23\textwidth]{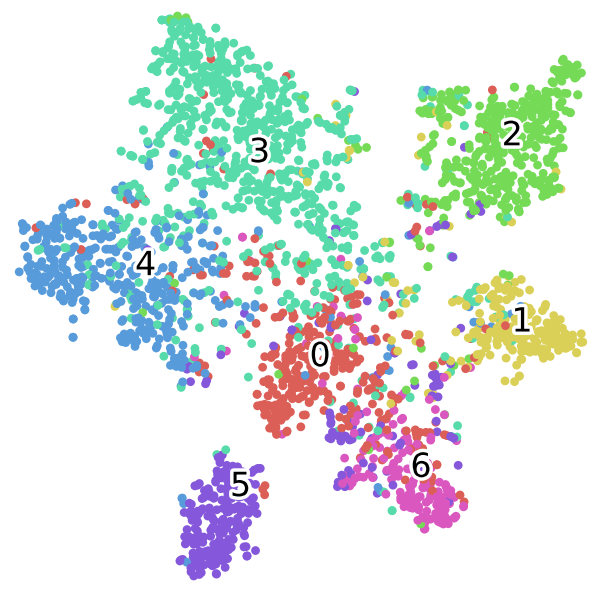}}
    \subfloat[GHNet(o)]{%
    \label{outer_cora}
    \includegraphics[width=0.23\textwidth]{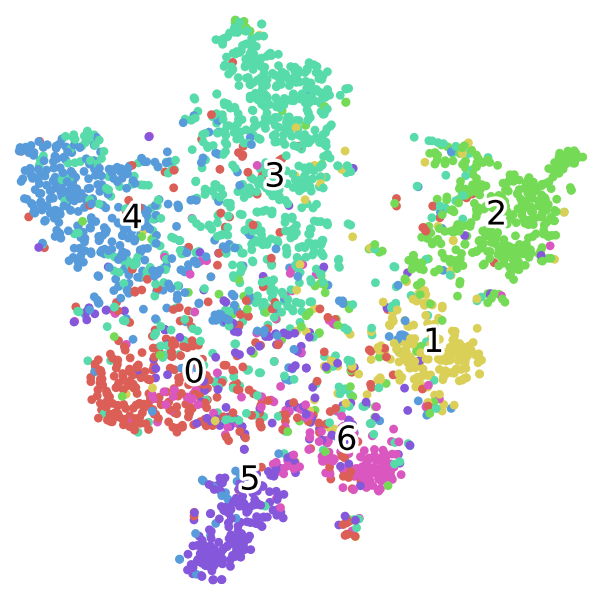}}
    \subfloat[GHNet(r)]{%
    \label{raw_cora}
    \includegraphics[width=0.23\textwidth]{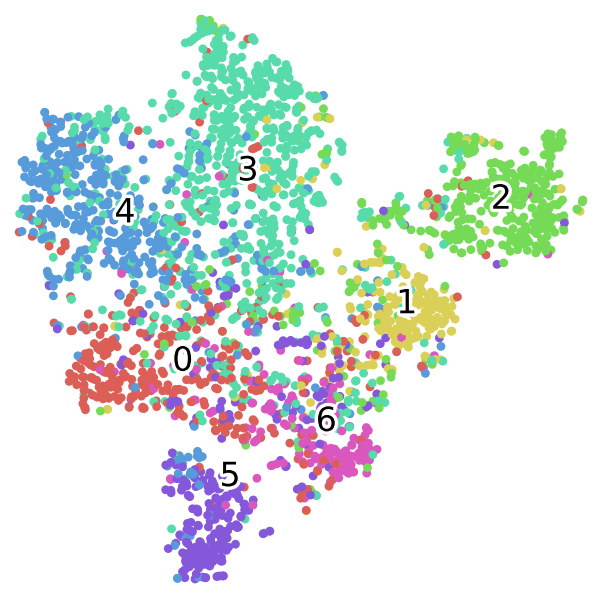}}
    \caption{Node embedding visualization on Cora dataset .}
    \label{embedding_ghnet}
\end{figure*}

To show whether GHNet resolves the over-smoothing problem, we visualize the learned node embeddings of the proposed three variants on Cora dataset. We set $k=5$ in the first convolution block and visualize this block's output. So the nodes now have a same receptive filed with 5-layer GCN.
Figure \ref{embedding_ghnet} shows the comparison of the node distribution. We can see that all three variants of GHNet don't suffer from the over-smoothing problem like the 5-layer GCN. The nodes belonging to the same class don't converge to alike vectors and still preserve their distinct information. Besides, different classes are also separated to different areas. This verifies the effectiveness of the proposed gating unites which can perform adaptive heterogeneity and homogeneity infusion.

\section{Related Work}
Learning from graph-structured data plays an important role in practical use cases. Plenty of research has been done to generalize neural networks to arbitrarily structured graphs \citep{duvenaud2015convolutional,bruna2013spectral,henaff2015deep,defferrard2016convolutional,li2015gatedgnn}. The first generation of graph convolutions is motivated from the spectral perspective \citep{bruna2013spectral,henaff2015deep,defferrard2016convolutional}. Based on this research, \citet{kipf2016GCN} proposed to use graph convolution networks for semi-supervised classification, now known as the famous GCN. Due to the success of GCN, both in terms of performance and computational complexity, there has been an abundance of research on the topic. \citet{rahimi2018geolocation} proposed to use GCN for user geolocation. \citet{berg2017graphmatrixcompletion} proposed to utilize GCN to perform matrix completion. \citet{kapoor2019mixhop} improved GCN by performing different hops of neighborhood mixing. \citet{graphsage} and \citet{chen2018fastgcn} proposed the sampling-based neighbourhood aggregation. \citet{wu2019SGC} treated the feature propagation as a pre-computing process and accelerated the learning of GCN. \cite{velivckovic2017GAT} proposed to use the attention mechanism other than topology (degree) information to learn the node importance when perform information propagation.

This work is broadly inspired by the observation that deeper architectures enabled by highway networks or residual connections have better expressiveness and fidelity in the deep learning field \citep{srivastava2015highway,srivastava2015trainingverydeep, he2016deep_residual_learning,he2016identity_mapping}. 
Some work was also conducted to investigate the depth problem of GCN, e.g. in the appendix of \citet{kipf2016GCN}, the authors conducted experiments to investigate the effectiveness of naive residual connections. However, the best performance was still obtained with shallow networks. \citet{li2019cangcngodeep} combined the dilation and residual connection and achieved a better performance in the task of point cloud segmentation. \citet{li2018deeper_insight} analyzed the over-smoothing problem from a spectral perspective and proposed to learn GCN with co-training and self-training approaches. \citet{xu2018jumping} proposed an ensemble learning approach that combines the hidden representations of different layers to enable better structure-aware representations. \citet{li2015gatedgnn} proposed the use of shared parameters and LSTM to build graph neural networks.

\section{Conclusion}
In this paper, we investigate the over-smoothing problem of GCN. We argue that the re-normalization trick used in GCN will lead to overly homogeneous information propagation and thus create the over-smoothing problem. We further state that there is a trade-off between homogeneity and heterogeneity in the learning procedure of GCN. Overly homogeneous and heterogeneous neighborhood aggregation will both affect the model performance. To automatically balance this trade-off, we propose GHNet which are featured with multi-hop feature propagation and learnable gating units. The former enables GHNet to achieve much larger receptive filed with small number of parameters while the later serves as an information highway to perform heterogeneous infusion so the node can preserve its own identical features. We conduct extensive experiments on benchmark datasets to verify our propositions. The results demonstrate that the proposed GHNet achieve superior performance than GCN and other related models. Future work including investigating the performance of GHNet on other application scenarios such as recommendation and genome annotation. Besides, we are also interested in generalizing GHNet to distributed environment and heterogeneous graphs.
\newpage

\bibliography{bibliography.bib}
\bibliographystyle{icml2020}





\end{document}